\documentclass[10pt,twocolumn,letterpaper]{article}

\usepackage{cvpr}              

\usepackage[dvipsnames]{xcolor}

\usepackage{graphicx}
\usepackage{amsmath}
\usepackage{amssymb}
\usepackage{booktabs}
\usepackage[export]{adjustbox}

\usepackage{placeins}
\usepackage{graphbox}
\usepackage{breqn}
\usepackage{multicol}
\usepackage{multirow}
 \usepackage{mathtools}
 \usepackage{soul}
\usepackage{makecell}
\usepackage{color, colortbl}
\usepackage{dsfont}
\usepackage{caption}
\usepackage{cite}
\usepackage{afterpage}

\newcommand{\expec}{\mathbb{E}}

\newcommand{\lsimple}{L_{DM}}
\newcommand{\lsimpleldm}{L_{LDM}}

\newcommand{\model}{\epsilon_\theta}

\definecolor{cvprblue}{rgb}{0.21,0.49,0.74}
\usepackage[pagebackref,breaklinks,colorlinks,citecolor=cvprblue]{hyperref}

\newcommand{\eref}[1]{(\ref{#1})}

\newcommand{\tabref}[1]{Table~\ref{#1}}

\newcommand{\figref}[1]{Figure~\ref{#1}}

\newcolumntype{L}[1]{>{\raggedright\let\newline\\\arraybackslash\hspace{0pt}}m{#1}}
\newcolumntype{C}[1]{>{\centering\let\newline\\\arraybackslash\hspace{0pt}}m{#1}}
\newcolumntype{R}[1]{>{\raggedleft\let\newline\\\arraybackslash\hspace{0pt}}m{#1}}

\newcommand{\ourmethod}{FineControlNet}

\def\paperID{16811} 
\def\confName{CVPR}
\def\confYear{2024}

\title{FineControlNet: Fine-level Text Control for Image Generation with\\Spatially Aligned Text Control Injection}

\author{Hongsuk Choi\textsuperscript{*}, Isaac Kasahara\textsuperscript{*}, Selim Engin,  Moritz Graule, Nikhil Chavan-Dafle, and Volkan Isler\\
(* Indicates equal contributions)\\
Samsung AI Center, New York \\
}

\begin{document}

\twocolumn[{
		\renewcommand\twocolumn[1][]{#1}
		\vspace{-0.5em}
		\maketitle
		\vspace{-1.5em}
		\begin{center}
			\centering
			\includegraphics[width=\linewidth]{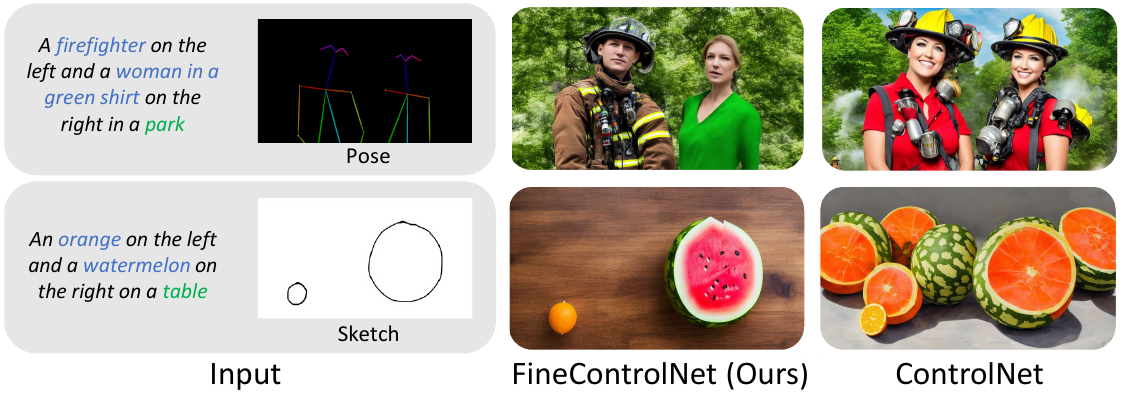}
			\vspace{-1.5em}
			\captionof{figure}{Our method, \ourmethod{}, generates images adhering to the user specified \textcolor{blue}{identities} and \textcolor{ForestGreen}{setting}, while maintaining the geometric constraints. Existing methods, such as ControlNet, merge or ignore the appearance and identity specified in the prompt.}
			\label{fig:front_teaser}
		\end{center}
}]

\maketitle

\begin{abstract}
Recently introduced ControlNet has the ability to steer the text-driven image generation process with geometric input such as human 2D pose, or edge features. 
While ControlNet provides control over the geometric form of the instances in the generated image, it lacks the capability to dictate the visual appearance of each instance. 
We present \ourmethod{} to provide fine control over each instance's appearance while maintaining the precise pose control capability. Specifically, we develop and demonstrate \ourmethod{} with geometric control via human pose images and appearance control via instance-level text prompts.
The spatial alignment of instance-specific text prompts and 2D poses in latent space enables the fine control capabilities of \ourmethod.
We evaluate the performance of \ourmethod{} with rigorous comparison against state-of-the-art pose-conditioned text-to-image diffusion models. \ourmethod{} achieves superior performance in generating images that follow the user-provided instance-specific text prompts and poses.

\end{abstract}

\section{Introduction}
Text-to-image diffusion models have become a popular area of research. With the release of production-ready models such as DALL-E 2~\cite{ramesh2022hierarchical} and Stable Diffusion~\cite{rombach2022high,sd2022}, users are able to generate images conditioned on text that describes characteristics and details of instances and background. 
ControlNet~\cite{zhang2023adding} enabled finer grained spatial control (i.e., pixel-level specification) of these text-to-image models without re-training the large diffusion models on task-specific training data.
It preserves the quality and capabilities of the large production-ready models by injecting a condition embedding from a separately trained encoder into the frozen large models.

While these models can incorporate the input text description at the scene level, the user cannot control the generated image at the object instance level.
%
For example, when prompted to generate a cohesive image with a person of specific visual appearance/identity on the left and a person of a different appearance/identity on the right, these models show two typical failures. Either one of the specified descriptions is assigned to both the persons in the generated image, or the generated persons show visual features which appear as interpolation of both the specified descriptions. Both of these peculiar failure modes can be seen in \figref{fig:front_teaser}. This lack of text-driven fine instance-level control limits the flexibility a user has while generating an image.

We address this limitation by presenting a method, called \ourmethod{}, that enables instance-level text conditioning, along with the finer grained spatial control (e.g. human pose). Specifically, we develop our method in the context of text-to-image generation with human poses as control input. However, note that the approach itself is not limited to this particular control input as we demonstrate in our supplementary materials.

Given a list of paired human poses and appearance/identity prompts for each human instance, \ourmethod{} generates cohesive scenes with humans with distinct text-specified identities in specific poses.
The pairing of appearance prompts and the human poses is feasible via large language models~\cite{chatgpt2022,touvron2023llama} (LLMs) or direct instance-specific input from the user.
The paired prompts and poses are fed to \ourmethod{} that spatially aligns the instance-level text prompts to the poses in latent space.



\begin{figure}[!t]
\begin{center}
\includegraphics[width=1.0\linewidth]{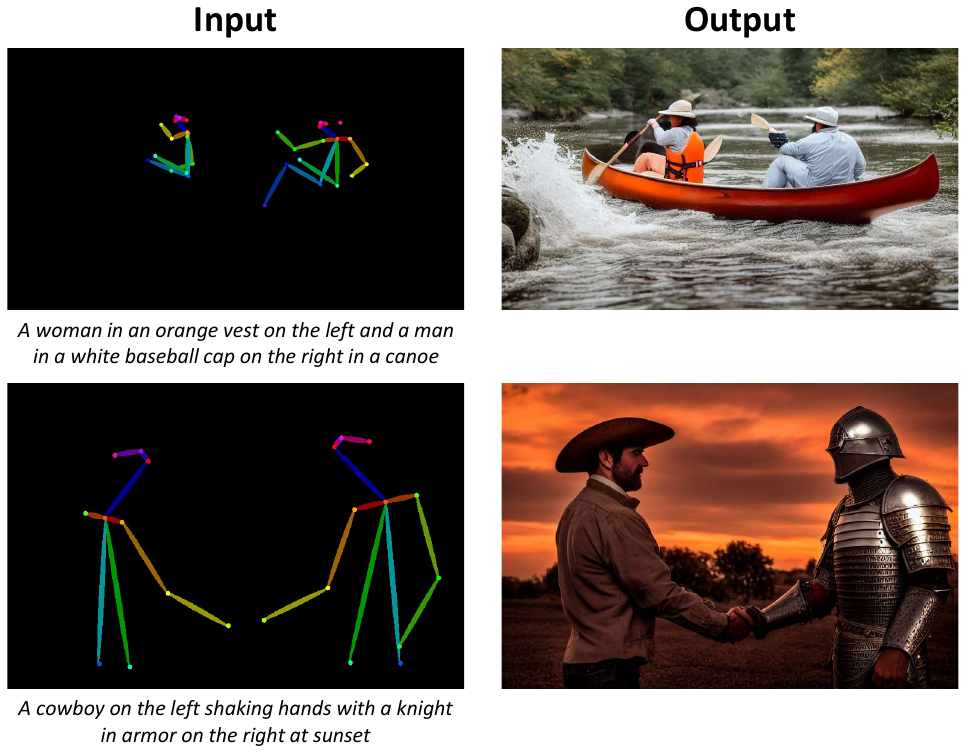}
\end{center}
\vspace{-2mm}
\caption{
\small{
Our \ourmethod{} generates images that ensure natural interaction between instances and environments, while preserving the specified appearance/identity and pose of each instance.
}
}
\label{fig:interaction}
\vspace{-2mm}
\end{figure}

\ourmethod{} is a training-free method that inherits the capabilities of the production-ready large diffusion models and is run in end-to-end fashion.
\ourmethod{} works by careful separation and composition of different conditions in the reverse diffusion (denoising) process.
In the initial denoising step, the complete noise image is copied by the number of instances.
Then, the noise images are processed by conditioning on separate pairs of text and pose controls in parallel, using the frozen Stable Diffusion and ControlNet.
During the series of cross attention operations in Stable Diffusion's UNet, the embeddings are composited using masks generated from the input poses and copied again.
This is repeated for every denoising step in the reverse diffusion process.
Through this latent space-level separation and composition of multiple spatial conditions, we can generate images that are finely conditioned, both in text and poses, and harmonize well between instances and environment as shown in \figref{fig:interaction}.

To evaluate our method, we compare against the state-of-the-art models for text and pose conditioned image generation. We demonstrate that \ourmethod{} achieves superior performance in generating images that follow the user-provided instance-specific text prompts and poses compared with the baselines. 

Our key contributions are as follows:

\begin{itemize}

    \item We introduce a novel method, \ourmethod{}, which gives a user the ability to finely control the image generation. 
    It fine controls generation of each instance in a cohesive scene context, using instance-specific geometric constraints and text prompts that describe distinct visual appearances. 

    \item We create a curated dataset and propose new metrics that focus on the evaluation of fine-level text control on image generation. The curated dataset contains 1000+ images containing multiple $(2 ~\textnormal{to}~ 15)$ humans in different poses per scene from the MSCOCO dataset~\cite{lin2015microsoft}. We label each human pose with appearance/identity description that will be used for generation, along with a setting description of the scene.
    

    \item Finally, we demonstrate our method's ability to provide the fine-grained text control against extensive state-of-the-art baselines. Our \ourmethod{} shows over 1.5 times higher text-image consistency metrics for distinguished multiple human image generation. Furthermore, we provide comprehensive qualitative results that support the robustness of our method.

\end{itemize}

\section{Related work}

\noindent\textbf{Text-to-Image Models:}
Text-to-image models have become a major area of research in the computer vision community. AlignDRAW~\cite{mansimov2015generating} introduced one of the first models to produce images conditioned on text. The field gained significant traction with the release of the visual-language model CLIP~\cite{radford2021learning} along with image generation model DALL-E~\cite{ramesh2021zero}. 
Diffusion models~\cite{sohl2015deep,ho2020denoising,song2020denoising} became the design of choice for text-to-image generation, starting with CLIP guided diffusion program~\cite{clipgdiff@crowson}, Imagen~\cite{saharia2022photorealistic}, and GLIDE~\cite{nichol2021glide}.
Latent Diffusion Model~\cite{rombach2022high}, DALL-E 2~\cite{ramesh2022hierarchical}, and VQ-Diffusion~\cite{gu2022vector} performed the diffusion process in latent space which has become popular for computational efficiency.
Due to the stunning quality of image generation, large diffusion models trained on large-scale Internet data~\cite{ramesh2022hierarchical,sd2022,midj2022} are now available as commercial products as well as sophisticated open-sourced tools.

\noindent\textbf{Layout- and Pose-conditioned Models:}
Generating images based on input types besides text is a popular concurrent research topic. 
Initial results from conditional generative adversarial network (GAN) showed great potential for downstream applications. Pix2Pix~\cite{isola2017image} could take in sketches, segmentation maps, or other modalities and transform them into realistic images.
LostGAN~\cite{sun2019image} proposed layout- and style-based GANs that enable controllable image synthesis with bounding boxes and category labels.
Recently, as text-to-image diffusion models became more popular, models that could condition both on text and other geometric inputs were studied. 
LMD: LLM-grounded Diffusion~\cite{lian2023llm}, MultiDiffusion~\cite{bar2023multidiffusion}, eDiff-I~\cite{balaji2022ediffi}, and GLIGEN~\cite{li2023gligen} condition the image generation on text as well as other 2D layout modalities such as segmentation masks. 
This allowed for text-to-image models to generate instances in specific areas of the image, providing more control to the user.

However, these methods are limited to positional control of individual instances, and did not extend to semantically more complex but spatially sparse modalities such as keypoints (e.g. human \textit{pose}).
For example, human pose involves more information than positional information, such as action and interaction with environment.
GLIGEN showed keypoints grounded generation, but it does not support multiple instances' pose control along with instance-specific text prompts. 
We analyze that it is mainly due to the concatenation approach for control injection~\cite{li2023gligen} that cannot explicitly spatially align the text and pose embeddings as in our method.


Recently, ControlNet~\cite{zhang2023adding} allowed the user to condition Stable Diffusion~\cite{sd2022} on both text as well as either segmentation, sketches, edges, depth/normal maps, and human pose without corrupting the parameters of Stable Diffusion. 
Along with the concurrent work T2I~\cite{mou2023t2i}, ControlNet sparked a large interest in the area of pose-based text-to-image models, due to their high quality of generated human images. 
HumanSD~\cite{ju2023humansd} proposed to fine-tune the Stable Diffusion using a heatmap-guided denoising loss.
UniControl~\cite{qin2023unicontrol} and DiffBlender~\cite{kim2023diffblender} unified the separate control encoders to a single encoder that can handle different combinations of text and geometric modalities, including human pose.
While these methods, including ControlNet, produce high quality images of multiple humans in an image, they lack the capability to finely dictate what individual human should look like through a high level text description per human. 
To address this limitation, we introduce \ourmethod{}, a method for generation conditioned on text and poses that has the capability to create images that are true-to-the-prompt at the instance-level as well as harmonized with the overall background and context described in the prompt.

\section{\ourmethod}
\label{sec:method}
The motivation for FineControlNet is to provide users with text and 2D \textit{pose} control beyond position for individual instances (i.e. human) during image generation.
FineControlNet achieves this by spatially aligning the specific text embeddings with the corresponding instances' 2D poses.

\begin{figure*}[!t]
\begin{center}
\includegraphics[width=1.0\linewidth]{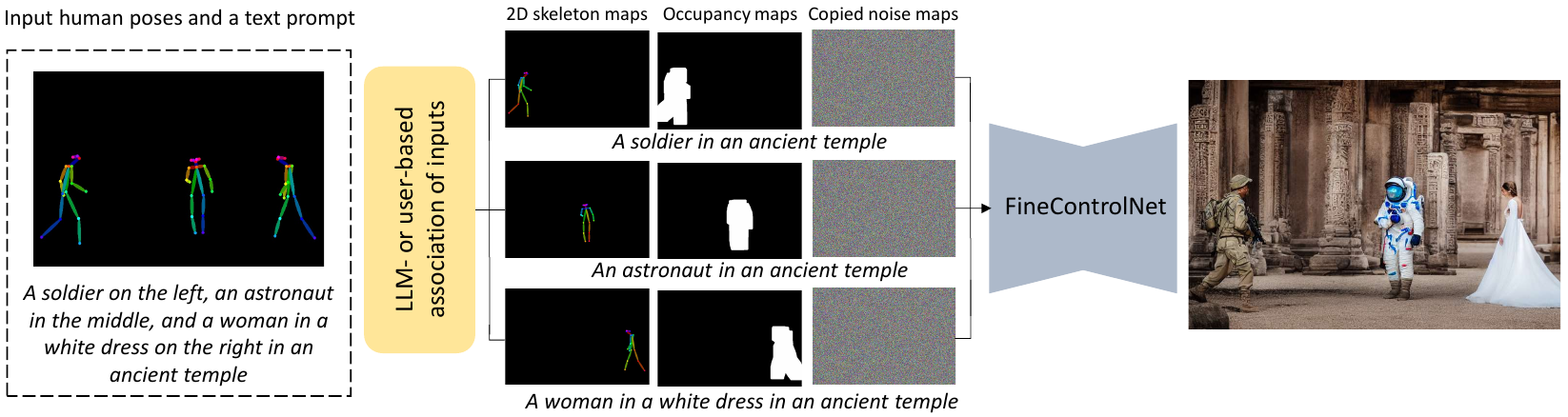}
\end{center}
\vspace{-3mm}
\caption{
\small{
\textbf{Method Overview.} Given a set of human poses as well as text prompts describing each instance in the image, we pass triplets of skeleton/mask/descriptions to \ourmethod{}. 
By separately conditioning different parts of the image, we can accurately represent the prompt's description of the appearance details, relative location and pose of each person.
}
}
\label{fig:FineControlNet_pipeline}
\vspace{-4mm}
\end{figure*}

\begin{figure}[!t]
\begin{center}
\includegraphics[width=1.0\linewidth]{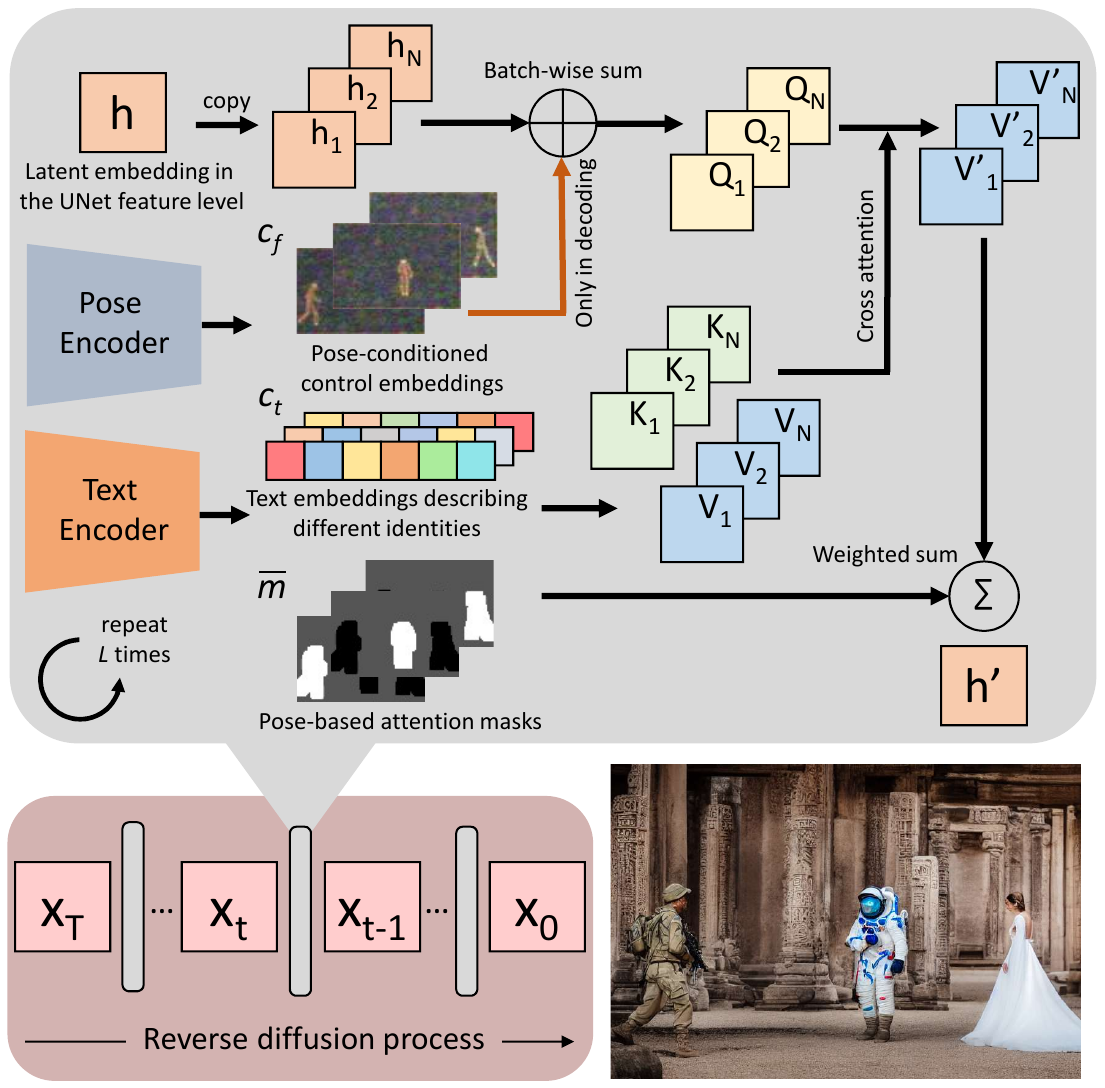}
\end{center}
\caption{
\small{
During the reverse diffusion process, \ourmethod{} performs composition of different instances' embeddings at each of the $L$ layers of UNet. Triplets of pose, text, and copy of the latent embedding $h$ are passed through each block of the UNet architecture in parallel.
The embeddings are composited after the cross attention using the normalized attention masks.
}
}
\label{fig:cross_attention}
\vspace{-4mm}
\end{figure}

\subsection{Preliminary}
\label{section:preliminary}
To better highlight our method and contributions, we first explain the key ideas of diffusion-based image generation.

Image generation using probabilistic diffusion models~\cite{nichol2021glide,ramesh2022hierarchical} is done by sampling the learned distribution $p_\theta(x_0)$ that approximates the real data distribution $q(x_0)$, where $\theta$ is learnable parameters of denoising autoencoders 
$\model(x)$. 
During training, the diffusion models gradually add noise to the image $x_0$ and produce a noisy image $x_t$.
The time step $t$ is the number of times noise is added and uniformly sampled from $\{1, \dots, T\}$.
The parameters $\theta$ are guided to predict the added noise with the loss function

\begin{equation}
\lsimple = \expec_{x, \epsilon \sim \mathcal{N}(0, 1),  t }\Big[ \Vert \epsilon - \model(x_{t},t) \Vert_{2}^{2}\Big] \, .
\label{eq:dmloss}
\end{equation}

During inference, the sampling (i.e. reverse diffusion) is approximated by denoising the randomly sampled Gaussian noise $x_T$ to the image $x_0$ using the trained network $\model(x)$.

Conditional image generation is feasible via modeling conditional distributions as a form of $p_\theta(x_0|c)$, where $c$ is the conditional embedding that is processed from text or a task-specific modality.
Recent latent diffusion methods~\cite{rombach2022high,zhang2023adding} augment the UNet~\cite{ronneberger2015u}-based denoising autoencoders by applying cross attention between noisy image embedding $z_t$ and conditional embedding $c$.
The network parameters $\theta$ are supervised as below:

\begin{equation}
\lsimpleldm = \expec_{x, \epsilon \sim \mathcal{N}(0, 1),  t }\Big[ \Vert \epsilon - \model(z_{t},t,c_t,c_f) \Vert_{2}^{2}\Big] \, ,
\label{eq:ldmloss}
\end{equation}

where $c_t$ is a text embedding and $c_f$ is a task-specific embedding that is spatially aligned with an image in general.

\subsection{Spatial Alignment of Text and 2D Pose}

While the conditional image generation works reasonably well at a global level, it becomes challenging when fine control over each of the instances with text prompts is desired.
Since text is not a spatial modality that can be aligned with image, it is ambiguous to distribute the text embeddings to corresponding desired regions.

We formulate this text-driven fine control problem as spatially aligning instance-level text prompts to corresponding 2D geometry conditions (i.e. 2D poses).
Given a list of 2D poses $\{p^{\text{2D}}_i\}_1^N$, we create a list of attention masks $\{m_i\}_1^N$, where $N$ is the number of humans. 
We obtain these masks by extracting occupancy from 2D pose skeletons and dilating them with a kernel size of $H/8$, where $H$ is the height of the image.
The occupancy maps are normalized by softmax and become attention masks $\{m_i\}_1^N$, where sum of mask values add up to $1$ at every pixel.

We define the latent embedding $h$ at each time step $t$, which collectively refers to the outputs of UNet cross-attention blocks, as composition of multiple latent embeddings $\{h_i\}_1^N$:




\begin{equation}
h = \overline{m}_1 \cdot h_1 + \overline{m}_2 \cdot h_2 + \dots + \overline{m}_N \cdot h_N  \, ,
\label{eq:latent_composition}
\end{equation}

where $h_i$ embeds the $i$th instance's text condition in the encoding step and text and 2D pose conditions in the decoding step, and $\overline{m}_i$ is a resized attention mask. 
Now, $h$ contains spatially aligned text embeddings of multiple instances.
The detailed composition process is described in \figref{fig:cross_attention}.
It graphically depicts how we implement equation \eref{eq:latent_composition} in a UNet's cross attention layer for text and 2D pose control embeddings.
In both encoding and decoding stages of UNet, copied latent embeddings $\{h_i\}_1^N$ are conditioned on instance-level text embeddings $\{c_t\}_1^N$ by cross attention in parallel.
In the decoding stage of UNet, instance-level 2D pose control embeddings $\{c_f\}_1^N$ are added to the copied latent embeddings $\{h_i\}_1^N$ before the cross attention.

\begin{figure}[!t]
\begin{center}
\includegraphics[width=1.0\linewidth]{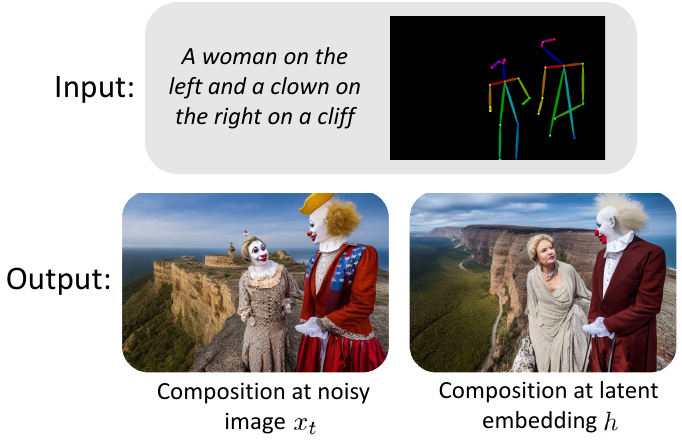}
\end{center}
\vspace{-5mm}
\caption{
\small{
Comparison between applying the composition step at the noise image $x$ vs. at the latent embedding $h$. Different from ours, visual features of instances are blended in a generated image due to the composition at $x$.
}
}
\label{fig:why_not_x}
\vspace{-4mm}
\end{figure}

Our composition of latent embeddings is inspired by the inpainting paper Repaint~\cite{lugmayr2022repaint}.
It is a training-free method and performs composition of known and unknown regions of noisy image $x_t$ similar to ours. 
However, the composition in the latent space level is fundamentally more stable for a generation purpose. 
In each DDIM~\cite{song2020denoising} step of the reverse diffusion, ${x_{t-1}}$ is conditioned on the predicted $x_0$ as below:

\begin{align}
    x_{t-1} = \sqrt{\alpha_{t-1}} x_0 + \sqrt{1 - \alpha_{t-1} - \sigma_t^2} \cdot \model(x_t) + \sigma_t \epsilon_t  \, , \\
    x_0 = \left(\frac{x_t - \sqrt{1 - \alpha_t} \model(x_t)}{\sqrt{\alpha_t}}\right) \, ,
\label{eq:sample-eq-gen}
\end{align}

where $\alpha_t$ is the noise variance at the time step $t$, $\sigma$ adjusts the stochastic property of the forward process, and $\epsilon_t$ is standard Gaussian noise independent of $x_t$.
As shown in the above equation, compositing multiple noisy images $x_{t-1}$ as that in inpainting literature is essentially targeting interpolation of multiple denoised images for generation.
On the contrary, the latent-level composition mathematically samples a unique solution from a latent embedding that encodes spatially separated text and pose conditions.
\figref{fig:why_not_x} supports the motivation of the latent-level composition.

\subsection{Implementation of FineControlNet}
\ourmethod{} is a training-free method that is built upon pretrained Stable Diffusion v1.5~\cite{sd2022} and ControlNet v1.1~\cite{zhang2023adding}.
We implemented our method using ControlNet’s pose-to-image model.
We modified its reverse diffusion process for fine-level text control of multiple people at inference time.
If a single pair of text and pose is given, our method is the same with ControlNet.
The whole process is run in an end-to-end fashion.
We do not require a pre-denoising stage to obtain fixed segmentation of instances as LMD~\cite{lian2023llm} nor inpainting as post-processing for harmonization.
The overall pipline of \ourmethod{} is depicted in \figref{fig:FineControlNet_pipeline}.

\noindent\textbf{Prompt parsing:} Our method requires instance level prompts that describe the expected appearance of each human. This differs slightly from competing methods, which generally only take in one prompt that describes the whole scene. While the user can manually prescribe each instance description to each skeleton (i.e. 2D pose) just as easily as writing a global prompt, large language models (LLM) can also be used as a pre-processing step in our method. If the user provides a global level description of the image containing descriptions and relative locations for each skeleton, many of the current LLM can take the global prompt and parse it into instance level prompts. Then given the center points of each human skeleton and the positioning location from the global prompt, an LLM could then assign each instance prompt to the corresponding skeleton. This automates the process and allows for a direct comparison of methods that take in detailed global prompts and methods that take in prompts per skeleton. An example of such processing is included in \figref{fig:FineControlNet_pipeline}.

\noindent\textbf{Harmony:}
We provide users with harmony parameters in addition to the default parameters of ControlNet.
Our text-driven fine control of instances has a moderate trade-off between identity instruction observance of each instance and the overall quality for image generation.
For example, if human instances are too close and the resolutions are low, it is more likely to suffer from \textit{identity blending} as ControlNet.
In such examples, users can decrease the softmax temperature of attention masks $\{m_i\}_1^N$, before normalization.
It will lead to better identity observance, but could cause discordant with surrounding pixels or hinder the denoising process due to unexpected discretization error in extreme cases.
Alternatively, users can keep the lower softmax temperature for initial DDIM steps and revert it back to a default value.
In practice, we use 0.001 as the default softmax temperature and apply argmax on the dilated pose occupancy maps for the first quarter of the entire DDIM steps.

\begin{table*}[!t]
\small
\centering
\setlength\tabcolsep{1.0pt}
\def\arraystretch{1.1}
\caption{\small{
Comparison with state-of-the-art pose and text-conditioned diffusion models. FineControlNet demonstrates superior scores in our CLIP Identity Observance (CIO) metrics  and is competitive in the rest of the metrics with the baselines. The CIO metrics measure the accuracy of each instance's appearance with relation to the prompt.}}
\vspace{-1mm}
\begin{tabular}{L{3.0cm}|C{2.2cm}|C{1.6cm}C{1.6cm}C{1.6cm}|C{1.5cm}C{1.5cm}C{1.5cm}|C{1.5cm}}
\specialrule{.1em}{.05em}{.05em}
\multirow{2}{*}{Methods} & \multicolumn{1}{c|}{Image Quality} & \multicolumn{3}{c|}{CLIP Identity Observance (CIO)} & \multicolumn{3}{c|}{Pose Control Accuracy} & \multirow{2}{*}{HND$\downarrow$} \\
& FID$\downarrow$ & $\text{CIO}_{\text{sim}}\uparrow$  & ${\text{CIO}_\sigma} \uparrow$ & $\text{CIO}_{\text{diff}}\uparrow$ & AP$\uparrow$ & $\text{AP}^M$$\uparrow$ & $\text{AP}^L$$\uparrow$ & \\ 
\specialrule{.1em}{.05em}{.05em}
ControlNet~\cite{zhang2023adding} & 5.85 & 23.0 & 0.34 & 1.4$\pm$1.5 & 56.1 & 10.2 & 60.6 & 5.4$\pm$6.0 \\
DiffBlender~\cite{kim2023diffblender} & 3.93 & 23.0 & 0.35 & 1.4$\pm$1.4 & 20.9 & 0.0 & 21.6 & 1.6$\pm$1.8 \\
GLIGEN~\cite{li2023gligen} & 4.02 & 22.2 & 0.34 & 1.0$\pm$1.2 & 64.9 & 5.4 & 67.5 & 1.9$\pm$2.6 \\
HumanSD~\cite{ju2023humansd} & 5.77 & 22.8 & 0.34 & 1.1$\pm$1.2 & 75.5 & 32.0 & 77.1 & 2.2$\pm$2.5 \\
UniControl~\cite{qin2023unicontrol} & 4.10 & 23.4 & 0.34 & 1.4$\pm$1.5 & 55.1 & 9.6 & 58.4 & 5.5$\pm$5.9 \\
T2I~\cite{mou2023t2i} & 10.30 & 23.1 & 0.34 & 1.4$\pm$1.5 & 58.3 & 14.1 & 62.1 & 5.8$\pm$6.5 \\ \hline
\ourmethod (Ours) & 4.05 & 24.2 & 0.56 & 2.9$\pm$2.3 & 63.2 & 16.7 & 65.9 & 2.4$\pm$3.1 \\ \hline
\specialrule{.1em}{.05em}{.05em}
\end{tabular}
\label{table:sota_table}
\vspace{-2mm}
\end{table*}

\section{Experiment}
In this section, we describe the set of experiments we conducted to assess the performance of \ourmethod{}.

\subsection{Setting}
\label{sec:setting}

\noindent\textbf{Baselines:}
For quantitative comparison, we chose the following state-of-the-art models in pose and text conditioned image generation: ControlNet~\cite{zhang2023adding}, UniControl~\cite{qin2023unicontrol}, HumanSD~\cite{ju2023humansd}, T2I~\cite{mou2023t2i}, GLIGEN~\cite{li2023gligen}, and DiffBlender~\cite{kim2023diffblender}.  These models allow for the user to specify multiple skeleton locations, as well as a global prompt.  
We convert the human poses in our MSCOCO-based dataset to Openpose~\cite{cao2019openpose} format for ControlNet, UniControl, and T2I, and convert to MMPose~\cite{mmpose2020} for HumanSD.  DiffBlender also allows the user to specify bounding boxes with additional prompts per bounding box.  For a fair comparison with our method, we input instance-level descriptions for bounding boxes around each skeleton in a similar fashion to our method.  DiffBlender and GLIGEN only support square inputs, so scenes are padded before being passed to these models, while the rest of the models take the images in their original aspect ratio. All models are used with their default parameters when running on our benchmark.


\noindent\textbf{Dataset:}
To evaluate the performance of our method and of the baselines at generating the images with multiple people using fine-grained text control, we introduce a curated dataset. This dataset contains over one thousand scenes with 2+ human poses per scene, extracted from the validation set of MSCOCO dataset~\cite{lin2015microsoft}. 
The total number of images is 1,126 and the total number of persons is 4,466. 

For the text annotation of the sampled data, we generated a pool of 50+ instance labels that describe a single person’s appearance, and a pool of 25+ settings that describe the context and background of the scene. We randomly assign each human pose an instance-level description, and each scene a setting description. We also include a global description that contains all of the instance descriptors with their relative positions described in text along with the setting descriptor.  This global description is used for baselines that can only be conditioned given a single prompt. 

\noindent\textbf{Metrics:}
The goal of our experiments is to evaluate our model's ability to generate cohesive and detailed images, generating pose-accurate humans, and allow for instance-level text control over the humans in the scene. 

We report the Fréchet Inception Distance (FID)~\cite{heusel2017gans} metric to measure the quality of the generated images, using the validation set of HumanArt~\cite{ju2023humanart} as the reference dataset. 

For measuring the text-image consistency between the generated image and the input text prompts, we introduce a set of new metrics called CLIP Identity Observance (CIO), based on CLIP~\cite{radford2021learning} similarity scores at the instance level. 
The first variant of this metric, $\text{CIO}_{\text{sim}}$, computes the similarity of text and image embeddings using the instance descriptions and the local patches around the input human poses.
While $\text{CIO}_{\text{sim}}$ evaluates the similarity of the generated image and the text prompt, it does not accurately measure the performance on synthesizing distinct identities for each instance. To address this limitation, we introduce $\text{CIO}_\sigma$ and $\text{CIO}_{\text{diff}}$ that measure how distinctly each instance description is generated in the image.

Given a local patch $I$ around an instance from the image and a set of text prompts $\mathcal{P}$ that describe all the instances, $\text{CIO}_\sigma$ computes a softmax-based score,

\begin{equation}
    \text{CIO}_\sigma = \frac{\exp\{\text{CLIP}(I, P^*)\}}{\sum_{P \in \mathcal{P}} \exp\{\text{CLIP}(I, P)\}}
\end{equation}

where $P^*$ is the text prompt corresponding to the instance in the local patch.
For the next metric, we compute the difference between the CLIP similarities of $P^*$ compared to text prompts describing other instances in the image, and define $\text{CIO}_{\text{diff}}$ as,

\begin{equation}
    \text{CIO}_{\text{diff}} = \text{CLIP}(I, P^*) - \sum_{P \in  \mathcal{P} - \{P^*\}} \frac{\text{CLIP}(I, P)}{|\mathcal{P}| - 1}
\end{equation}

To evaluate the pose control accuracy of methods, we test HigherHRNet~\cite{cheng2020higherhrnet} on our benchmark following HumanSD~\cite{ju2023humansd}.  
HigherHRNet is the state-of-the-art 2D pose estimator, and the weights are trained on MSCOCO and Human-Art~\cite{ju2023humanart} by the authors of HumanSD. 
We report the average precision (AP) of Object Keypoint Similarity (OKS)~\cite{lin2015microsoft} measured in different distance thresholds.
The superscript categorizes the resolution of people in an image and measures the average precision only for persons in that category, where $M$ and $L$ denote \textit{medium} and \textit{large} respectively.
Note that these metrics are pseudo metrics, because they are susceptible to inaccuracies of the 2D pose estimator independent from the inaccuracy of image generation.

\afterpage{\clearpage}
\begin{figure*}[p]
\begin{center}
\includegraphics[width=1.0\linewidth]{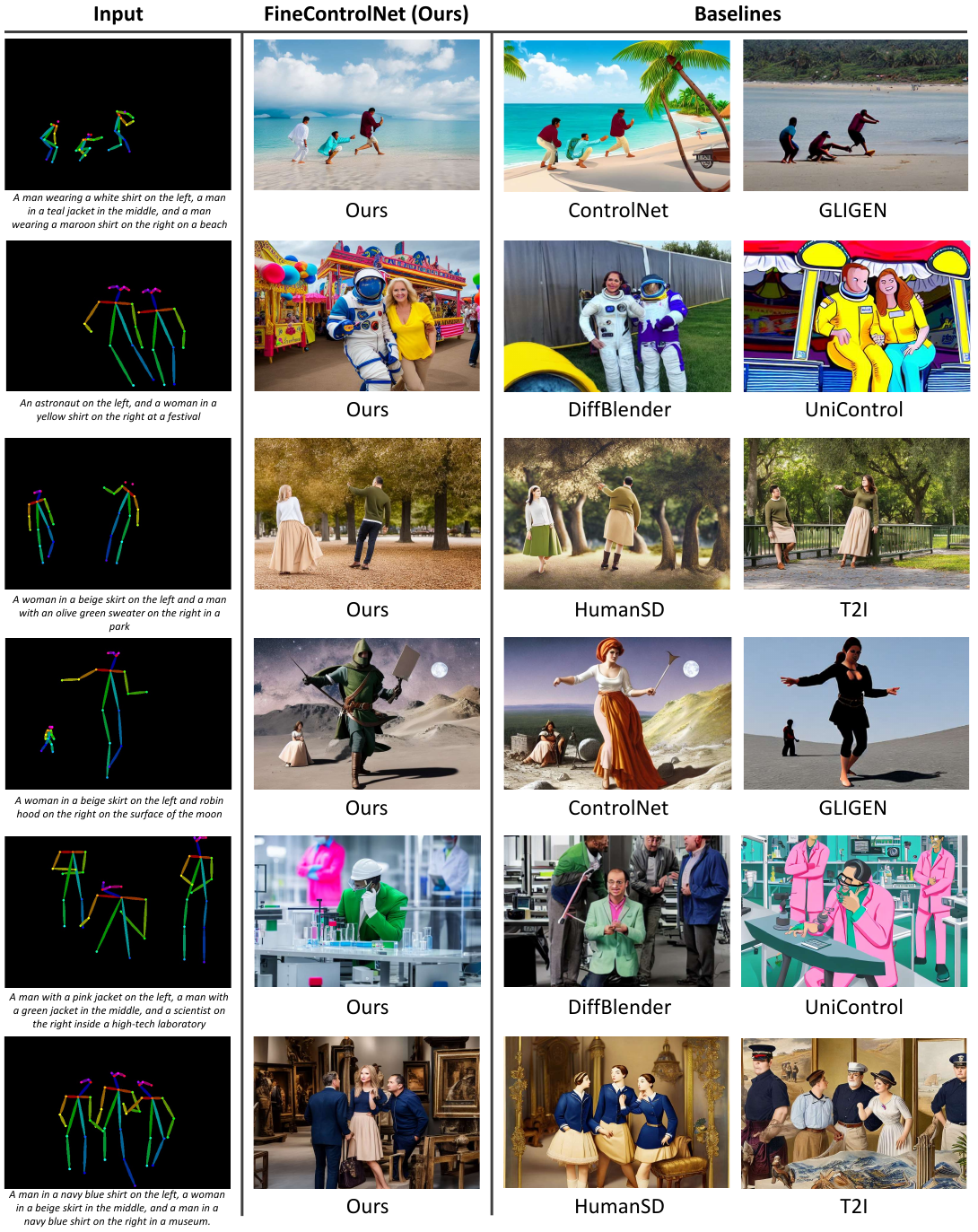}
\end{center}
\caption{
\small{
Qualitative results comparing FineControlNet against six state-of-the-art baselines. The input description and pose conditions are provided on the left side. FineControlNet consistently follows the user specified appearance prompts better in its image generations.}
}
\label{fig:qualitative_results}
\vspace{-2mm}
\end{figure*}

\subsection{Comparison with State-of-the-Art Methods}
We conduct quantitative analysis based on our benchmark and provide an extensive qualitative comparison.

\noindent\textbf{Quantitative Analysis:}
We evaluate methods on our benchmark dataset and assess the overall image quality, identity observance, and pseudo pose control accuracy of generated images using the metrics described in section~\ref{sec:setting}. Table~\ref{table:sota_table} shows the results from running our method as well as the six baselines.

We use FID metric as an overall measure of the quality that each method produces. We found that while DiffBlender and GLIGEN achieved the best  results in this category, our method is within the top half of the baselines.

For pose control accuracy evaluation, we report overall AP as well as AP with only medium-sized skeletons and only large-sized skeletons. Our method performs robustly in these categories, only behind HumanSD and comparable to GLIGEN. We believe the strong performance of HumanSD in APs is due to the the training bias of the tested HigherHRNet~\cite{cheng2020higherhrnet}.
The HigherHRNet is trained on Human-Art dataset for 2D pose estimation. HumanSD uses the same training dataset for image generation, while other methods are not explicitly trained on this dataset.

The Human Number Difference (HND) metric reports on average the difference between the number of ground truth skeletons in the input vs. the number of detected skeletons on the generated image. We find our method performs adequately here, outperforming ControlNet, T2I, and UniControl but underperforming against the other baselines.

The metric CLIP Identity Observance (CIO) is most relevant to the problem that we are targeting within this paper. This metric measures the instance description against the image patch that is supposed to match that description using a pretrained CLIP model~\cite{radford2021learning}. $\text{CIO}_{\text{sim}}$ does this directly, and we find that our method outperforms all the baselines in loyalty to the prompt. To further understand how much blending of different visual features happens for each methods outputs, we introduce the metrics $\text{CIO}_\sigma$ and $\text{CIO}_{\text{diff}}$. These metrics not only compare against the ground truth instance description, but also punish generated instances if they match other instances descriptions. For example if the instance label is ``Astronaut" and the other instance in the image is ``Soldier", the image will be punished if CLIP detects similarity between the image crop of an astronaut with the text ``Soldier". We found that our method strongly outperforms the baselines with these metrics, further demonstrating the effectiveness of using our reverse diffusion process conditioning individual instances only with their respective descriptions. We found that since DiffBlender is also given instance descriptions assigned to the skeleton locations in the form of bounding boxes, it came in second place overall in CIO. However it was not immune to blending features as seen in our qualitative results section.


\noindent\textbf{Qualitative Analysis:}
We generate a full page of qualitative results from our benchmark test in Figure~\ref{fig:qualitative_results} to demonstrate visually the difference between our method and the baselines. We also provide the input skeletons and prompt for the reader to compare how closely each method's images adhere to the description of the scene. We find that while all the methods can produce visually pleasing results and can incorporate the words from the description into the image, only ours is reliably capable of adhering to the prompt when it comes to the human poses/descriptions of each instance.

For example, ControlNet tends to blend or ignore features from all given text descriptions for example ignoring "robin hood" and only creating "woman in a beige skirt". GLIGEN also struggles to maintain identities and seems to average the different instance descriptions.

DiffBlender could sometimes distinguish instances into the correct locations, but would struggle adhering to both the pose condition and the text condition at the same time. Unicontrol appeared to focus on certain parts of the prompt, e.g. ``pink jacket", and use them to generate every instance.

Finally, HumanSD and T2I both suffer from the same issues of blending visual features between instances. Comparing the baselines to our method, we can see clear improvement in maintaining identity and characteristics from the input description.

\begin{table}[!t]
\small
\centering
\setlength\tabcolsep{1.0pt}
\def\arraystretch{1.1}
\caption{\small{
Ablation on level of composition. While our ablation using embedding $x$ produced higher pose scores and our ablation $h$-v2 produced higher CIO scores, we found that using embedding $h$ struck a good balance between pose and appearance control.}}
\vspace{-1mm}
\begin{tabular}{L{1.6cm}|C{1.4cm}|C{1.4cm}C{1.4cm}|C{1.4cm}}
\specialrule{.1em}{.05em}{.05em}
embedding & FID$\downarrow$ & ${\text{CIO}_\sigma}\uparrow$ & $\text{CIO}_{\text{diff}}\uparrow$ & AP$\uparrow$ \\
\specialrule{.1em}{.05em}{.05em}
$x$ & 2.86 & 0.35 & 1.39 & 68.0 \\
$h$-v2 & 3.36 & 0.60 & 3.20 & 45.7 \\ \hline
$h$ (Ours) & 4.05 & 0.56 & 2.91 & 63.2 \\ \hline
\specialrule{.1em}{.05em}{.05em}
\end{tabular}
\label{table:compose_level_ablation}
\vspace{-2mm}
\end{table}


\subsection{Ablation Study}
We study alternative ways of composition of different instances' conditions.
First, we perform the composition in the level of denoised image $x$.
Second, we modify the current composition in the level of latent embedding $h$, which we name $h$-v2 in~\tabref{table:compose_level_ablation}.
We apply the composition of different pose embeddings before the cross attention and add them to $\{h_i\}_1^N$.
The pre-composition before the cross attention is repeated for every decoding step of UNet.
The final output of the UNet is then composed using the attention masks $\{m_i\}_1^N$.

As shown in~\tabref{table:compose_level_ablation}, the composition of $x$ presents 37.5\% lower ${\text{CIO}_\sigma}$, while the pose control accuracy increases only by 7.9\% .
The quantitative results support our statement that it is essentially targeting interpolation of multiple denoised images for generation.
The results are also aligned with observation in~\figref{fig:why_not_x}.
Our modified composition in $h$ level shows the slightly better accuracy in CIO, but gives the worst pose control accuracy that is 27.7\% lower than ours.
We conjecture that the composition of pose embeddings prior to the cross attention weakens the individual instance's control signal after the attention, due to the distribution of attention to other instances' poses.



\section{Conclusion}

We introduced \ourmethod{}, a novel method to finely control instance level geometric constraints as well as appearance/identity details for text-to-image generation. Specifically, we demonstrated the application for generating images of humans in specific poses and of distinct appearances in a harmonious scene context. 
\ourmethod{} derives its strength from the spatial alignment of the instance-level text
prompts to the poses in latent space. During the reverse diffusion process, the repeated composition of embeddings of spatial-geometric and appearance-text descriptions leads to a final image that is conditioned on text and poses, and is consistent with the overall scene description.

To evaluate the performance of \ourmethod{} and comparable baselines for pose-conditioned text-to-image generation, we introduced a curated benchmark dataset based off of the MSCOCO dataset. With qualitative and quantitative analysis, we observed that \ourmethod{} demonstrated superior performance on instance-level text-driven control compared with the state-of-the-art baselines.
\ourmethod{} provides the enhanced control over the form and appearance for image generation, pushing the frontiers of text-to-image generation capabilities further.



\bibliographystyle{ieeenat_fullname}
\bibliography{main}

\clearpage

\twocolumn[{
\begin{center}
\begin{Large}
\textbf{\large Supplementary Material of\\FineControlNet: Fine-level Text Control for Image Generation with\\Spatially Aligned Text Control Injection}

\vspace{1.5em}

\centering
\includegraphics[width=1\linewidth]{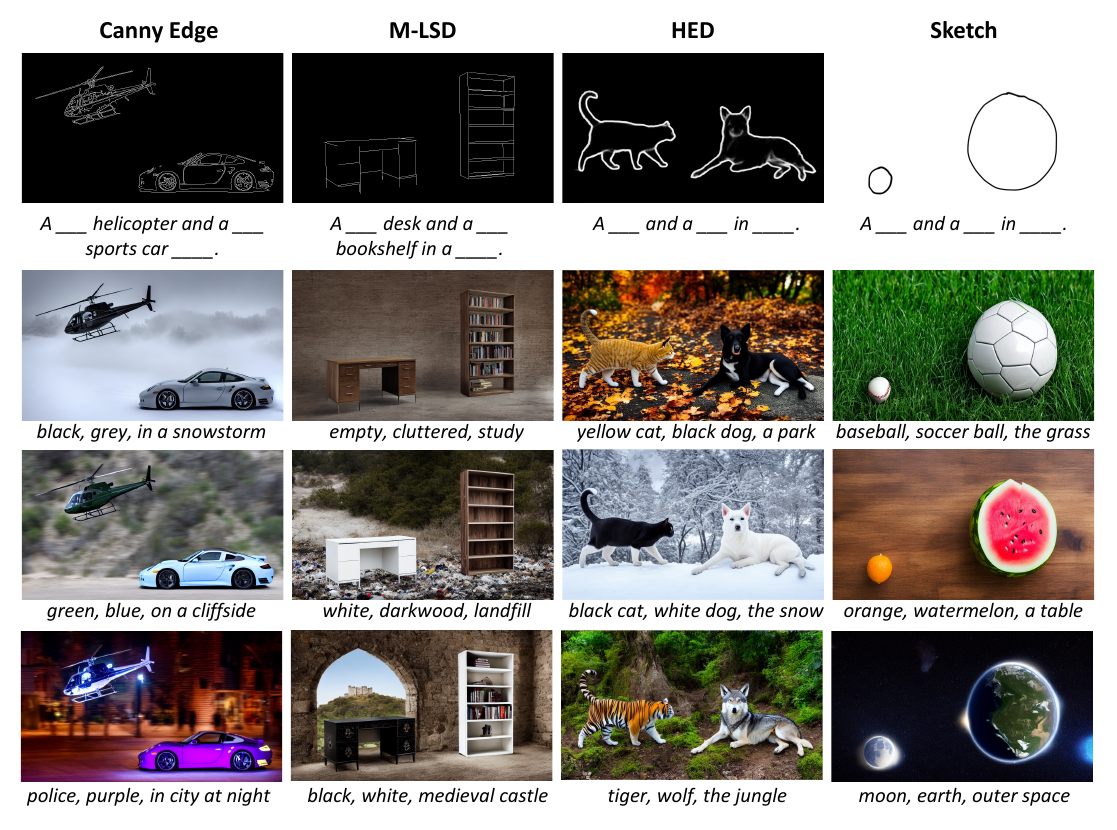}
\captionof{figure}{Results of \ourmethod{} applied to different control modalities of Canny~\cite{canny1986computational} edges, M-LSD~\cite{gu2022towards} lines, HED~\cite{xie2015holistically} edges, and a sketch input. As shown above our method has the ability to not only work on human pose inputs, but other modalities as well using the same approach described in our method section but applied to different ControlNet\cite{zhang2023adding} models. Under each column is the modality name, the sample input image, the prompt template, and three examples images with the corresponding input prompt information. Our method demonstrates the ability to finely control each instance.}
\label{fig:new_modality}
  
\end{Large}
\end{center}
\vspace*{+2em}
}]

In this supplementary material, we present more experimental results that could not be included in the main manuscript due to the lack of space. 

\begin{table*}[!t]
\small
\centering
\setlength\tabcolsep{1.0pt}
\def\arraystretch{1.1}
\caption{\small{
Robustness Study regarding factors of ``number of people'', ``scale of a person'', and ``distance between people''.}}
\vspace{-1mm}
\begin{tabular}{C{1.5cm}|C{1.2cm}C{1.2cm}C{1.2cm}|C{1.2cm}C{1.2cm}C{1.2cm}C{1.2cm}C{1.2cm}|C{1.2cm}C{1.2cm}C{1.2cm}C{1.2cm}}
\specialrule{.1em}{.05em}{.05em}
\multirow{2}{*}{Metrics} & \multicolumn{3}{c|}{Number of People} & \multicolumn{5}{c|}{Scale of a Person} & \multicolumn{4}{c}{Distance between People}\\
& 3 & 5 & 7 & 1 & 0.75 & 0.5 & 0.25 & 0.1 & 1 & 0.75 & 0.5 & 0.25\\
\specialrule{.1em}{.05em}{.05em}
$\text{CIO}_{\text{sim}}\uparrow$ & 28.2 & 26.9 & 26.5 & 28.2 & 27.5 & 26.4 & 23.2 & 20.3 & 28.2 & 27.8 & 27.8 & 25.3\\
${\text{CIO}_\sigma}$ & 0.74 & 0.46 & 0.32 & 0.74 & 0.69 & 0.62 & 0.55 & 0.42 & 0.74 & 0.7 & 0.69 & 0.48\\
$\text{CIO}_{\text{diff}}\uparrow$ & 5.3$\pm$2.4 & 3.2$\pm$1.9 & 2.2$\pm$1.3 & 5.3$\pm$2.4 & 4.6$\pm$2.5 & 3.6$\pm$1.9 & 2.0$\pm$1.4 & 0.9$\pm$0.7 & 5.3$\pm$2.4 & 4.8$\pm$2.3 & 4.6$\pm$2.6 & 2.2$\pm$1.3\\ \hline
\specialrule{.1em}{.05em}{.05em}
\end{tabular}
\label{table:robustness_study}
\end{table*}

\begin{figure*}[!t]
\begin{center}
\includegraphics[width=1.0\linewidth]{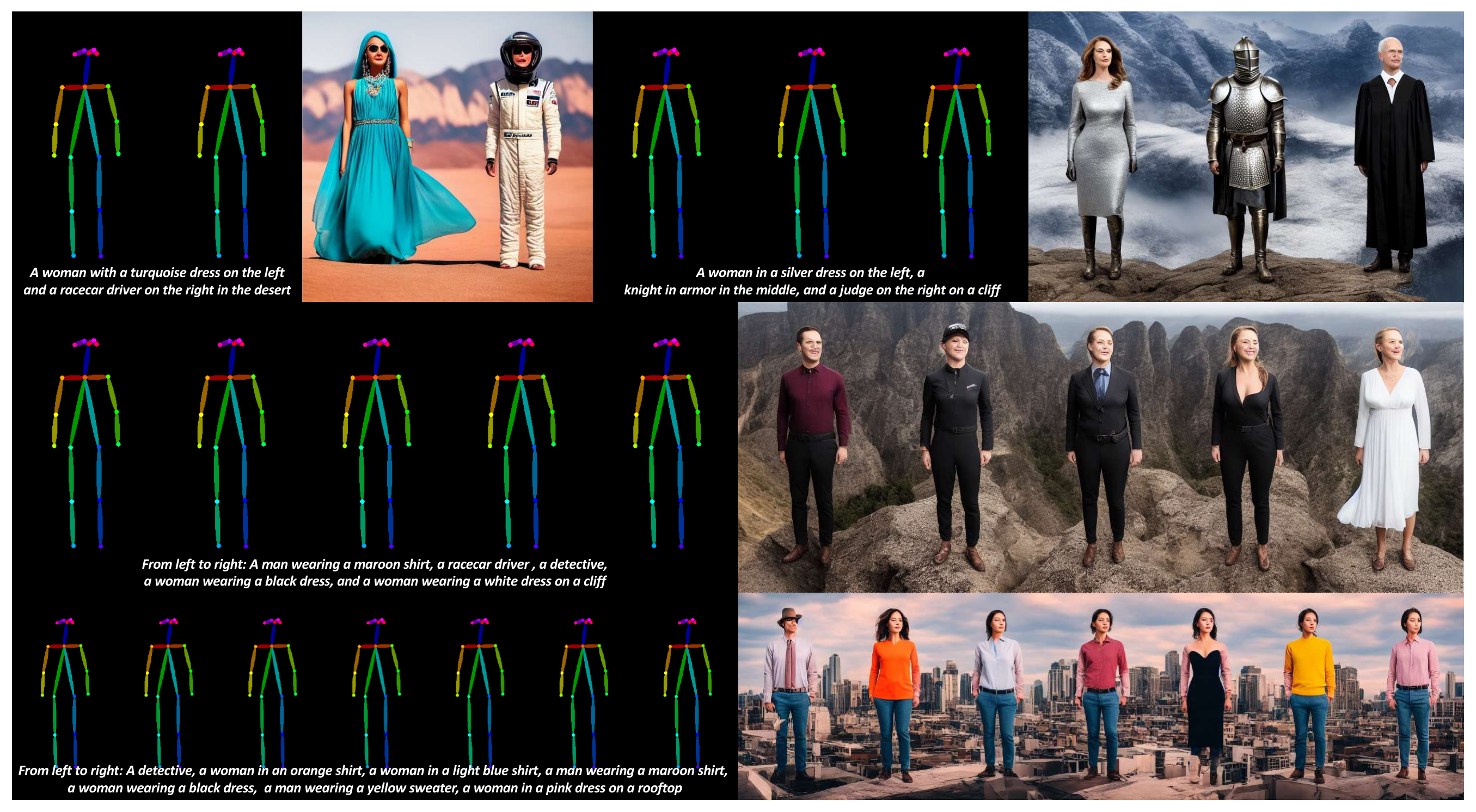}
\end{center}
\caption{
\small{
Qualitative results depending on the number of people, which is the number of 2D poses given. Every 2D human pose in the entire figure has the same resolution. The input skeleton map with 7 poses is resized to match the page. 
}
}
\label{fig:robustness_study_num_people}
\end{figure*}

\begin{figure*}[!t]
\begin{center}
\includegraphics[width=1.0\linewidth]{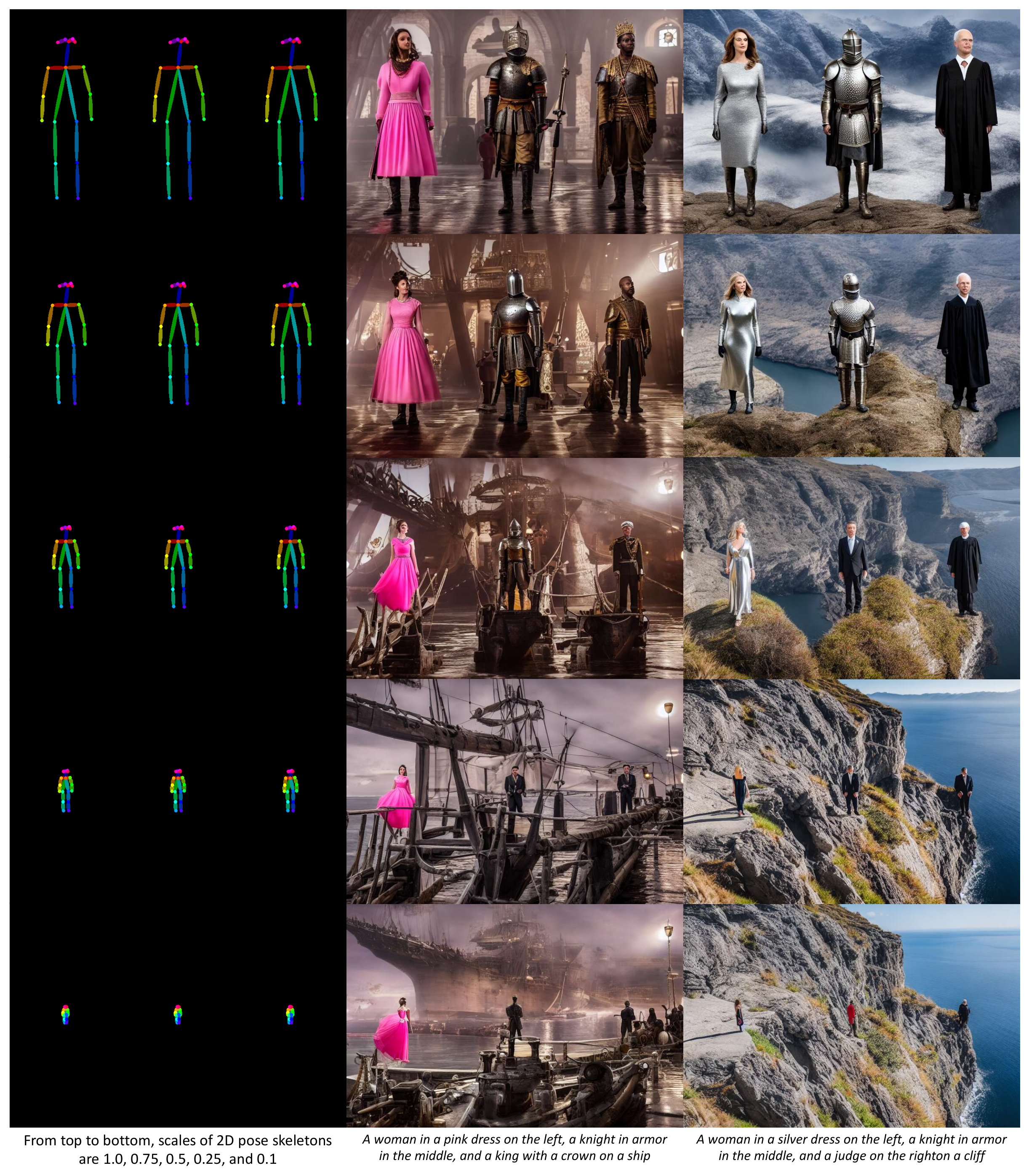}
\end{center}
\caption{
\small{
Qualitative results depending on the scale of a person, which represents the relative resolution of each pose in the input. We used the same seed for image generation for every scale variation. 
}
}
\label{fig:robustness_study_scale}
\end{figure*}

\begin{figure*}[!t]
\begin{center}
\includegraphics[width=1.0\linewidth]{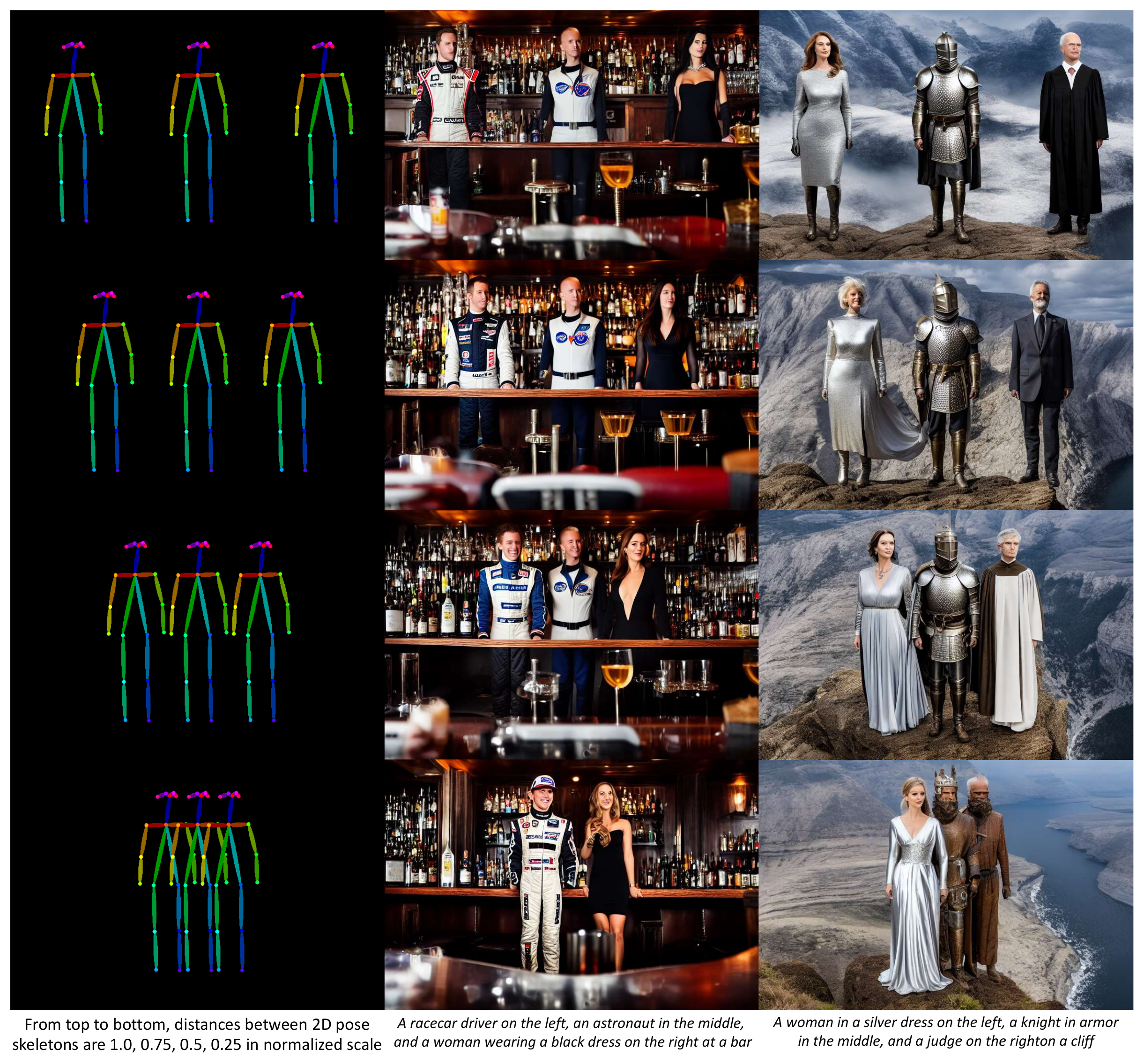}
\end{center}
\caption{
\small{
Qualitative results depending on the distance between people. Closer distance could cause \textit{blending} between different instances' text embeddings and generate mixed appearance of instances. We used the same seed for image generation for every inter-personal distance variation.
}
}
\label{fig:robustness_study_distance}
\end{figure*}

\section{Different Control Modality}
We present results demonstrating the efficacy of our \ourmethod{} architecture using various geometric control modalities, including Canny~\cite{canny1986computational} edges, M-LSD~\cite{gu2022towards} lines, HED~\cite{xie2015holistically} edges, and sketch inputs. As illustrated in \figref{fig:new_modality}, our framework enables fine-grained text-based control over individual instances while maintaining coherence across the generated scene. Through spatially aligned text injection, each instance faithfully reflects the corresponding textual prompt, with harmonized style and lighting that is consistent both within and between instances. For example, the bottom left image generated from the prompt “A police helicopter and a purple sports car in city at night” supports these claims; both vehicles exhibit glossy textures and lighting congruent with the nocturnal urban setting.


\section{How Robust is FineControlNet?}

We analyze the robustness of \ourmethod{} to variations in number of people, scale, and inter-personal distance. Quantitative experiments recording CLIP Identity Observance (CIO) scores (\tabref{table:robustness_study}) and qualitative results (Figures~\ref{fig:robustness_study_num_people}-\ref{fig:robustness_study_distance}) demonstrate performance under differing conditions.


Varying the number of input 2D poses while fixing scale and spacing reveals strong text-image consistency for 2-3 people, with gradual degradation as count increases to 5 and 7 (\tabref{table:robustness_study}; \figref{fig:robustness_study_num_people}). For instance, the fourth person from the left in \figref{fig:robustness_study_num_people} fails to wear the prompted dress, illustrating compromised identity observance. We posit that as instance count rises, pressure to balance identity adherence against holistic visual harmonization becomes more severe, increasing feature sharing between instances.


Experiments assessing robustness to variations in human scale utilize three input poses while fixing inter-personal distances. As depicted in~\figref{fig:robustness_study_scale} and~\tabref{table:robustness_study}, identity observance degrades gradually with increased downscaling, tied to losses in latent feature resolution. Performance remains reasonable down to 50\% scale, with more significant drops emerging under extreme miniaturization. Note input pose map resolution is constant at 512 pixels in height.


Similarly, distance experiments alter spacing around a central pose with three total people at fixed scale. Results in~\figref{fig:robustness_study_distance} and~\tabref{table:robustness_study} demonstrate consistent identity retention given non-overlapping inputs, with overlap introducing instance dropping or blending.


Together, these analyses quantify trade-offs between fidelity and spatial configurations. Performance gracefully handles reasonable perturbations but breaks down at data distribution extremes. Addressing such generalization limits highlights an area for further improvement.

\begin{figure}[t]
\begin{center}
\includegraphics[width=1.0\linewidth]{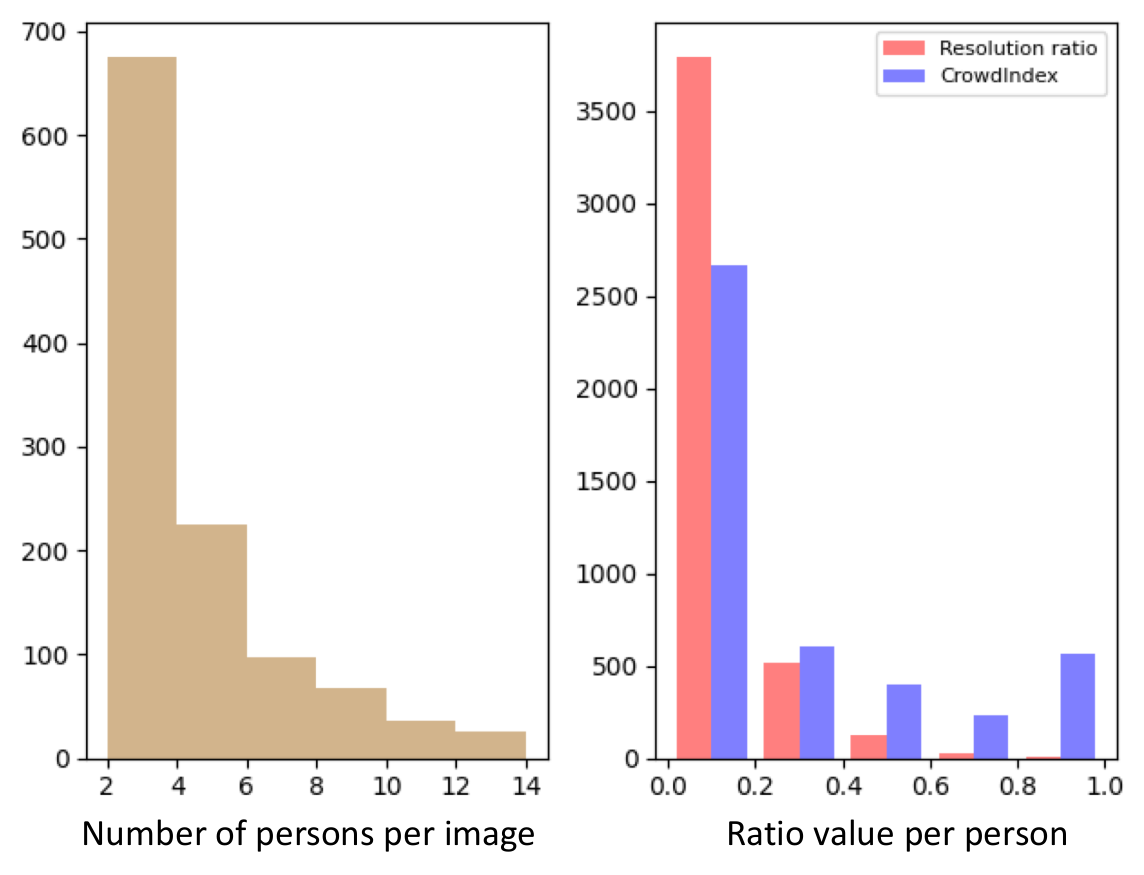}
\end{center}
\caption{
\small{
Statistics of our curated dataset. The y-axis indicates the counts that fall in bins in the x-axis.}
}
\label{fig:dataset_stats}
\vspace{-2mm}
\end{figure}

\section{Difference with MultiControlNet}
\begin{figure*}[!t]
\begin{center}
\includegraphics[width=1.0\linewidth]{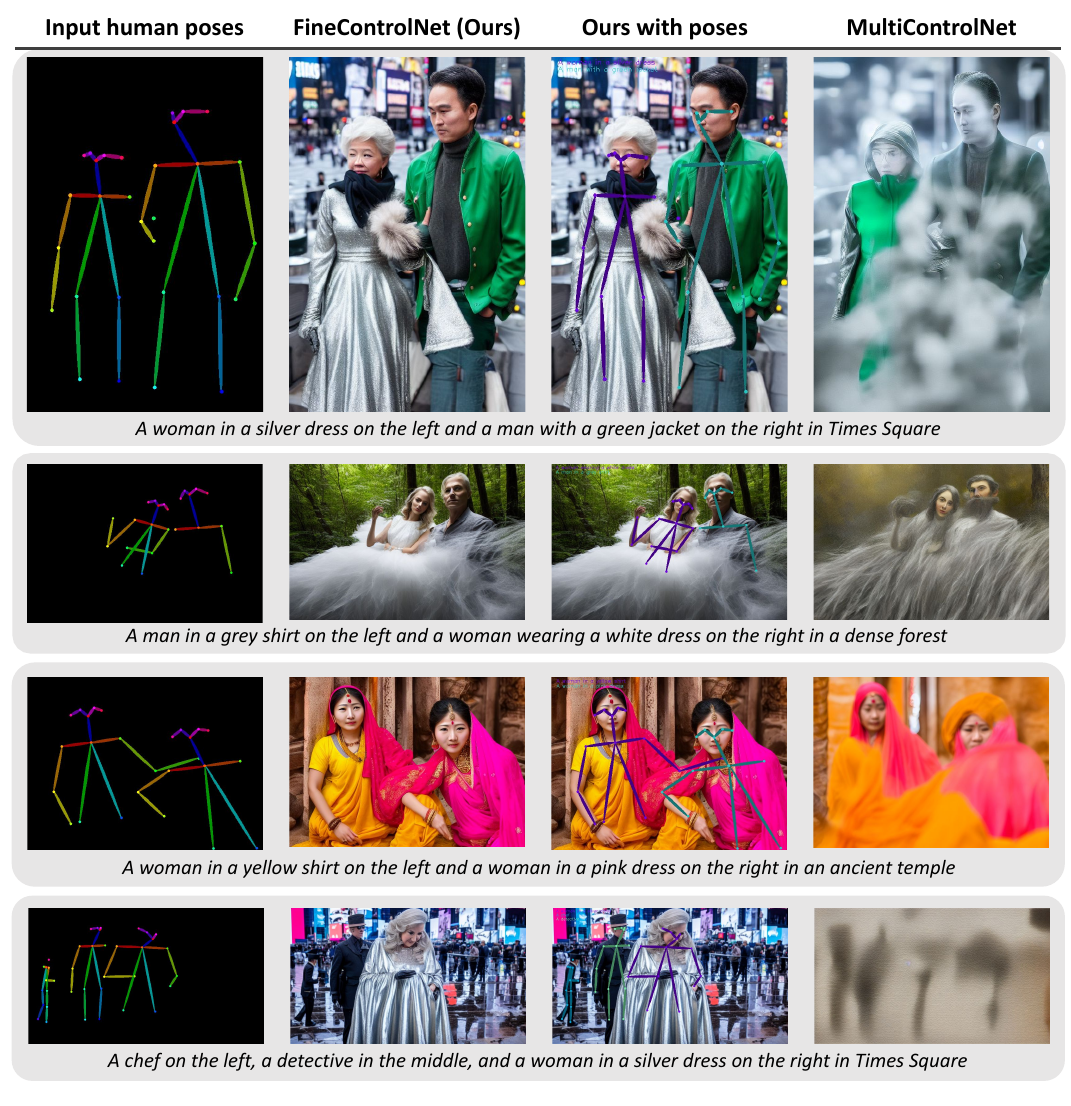}
\end{center}
\vspace{-3mm}
\caption{
\small{
Comparison between our \ourmethod{} and MultiControlNet~\cite{zhang2023adding,diffusers2023}. MultiControlNet produces blurry images, which also have blended appearance/identity between instances. In addition, more than two geometric control inputs paired with different text prompts often cause a complete failure. We provide the images of poses overlaid on \ourmethod's generated outputs for reference.
}
}
\label{fig:multicontrolnet_comparison}
\vspace{-4mm}
\end{figure*}

We compare \ourmethod{} to MultiControlNet~\cite{zhang2023adding}, an extension of ControlNet supporting multiple geometric modalities (e.g. pose, depth) with a single text prompt. For equivalence, we modify MultiControlNet to condition on instance-specific texts over multiple poses. Experiments utilize a third-party HuggingFace Diffusers~\cite{diffusers2023} implementation. Results in~\figref{fig:multicontrolnet_comparison} demonstrate compromised adherence to per-instance textual prompts compared to \ourmethod, stemming from lack of spatial text-latent alignment and careful latent composition. Moreover, MultiControlNet fails to process more than two inputs, generating blurry and abstract imagery. These contrasts highlight the importance of \ourmethod’s spatially aware text injection and carefully engineered latent fusion for fine-grained, multi-instance control.



\section{More Qualitative Results}

\begin{figure*}[p]
\begin{center}
\includegraphics[width=0.9\linewidth]{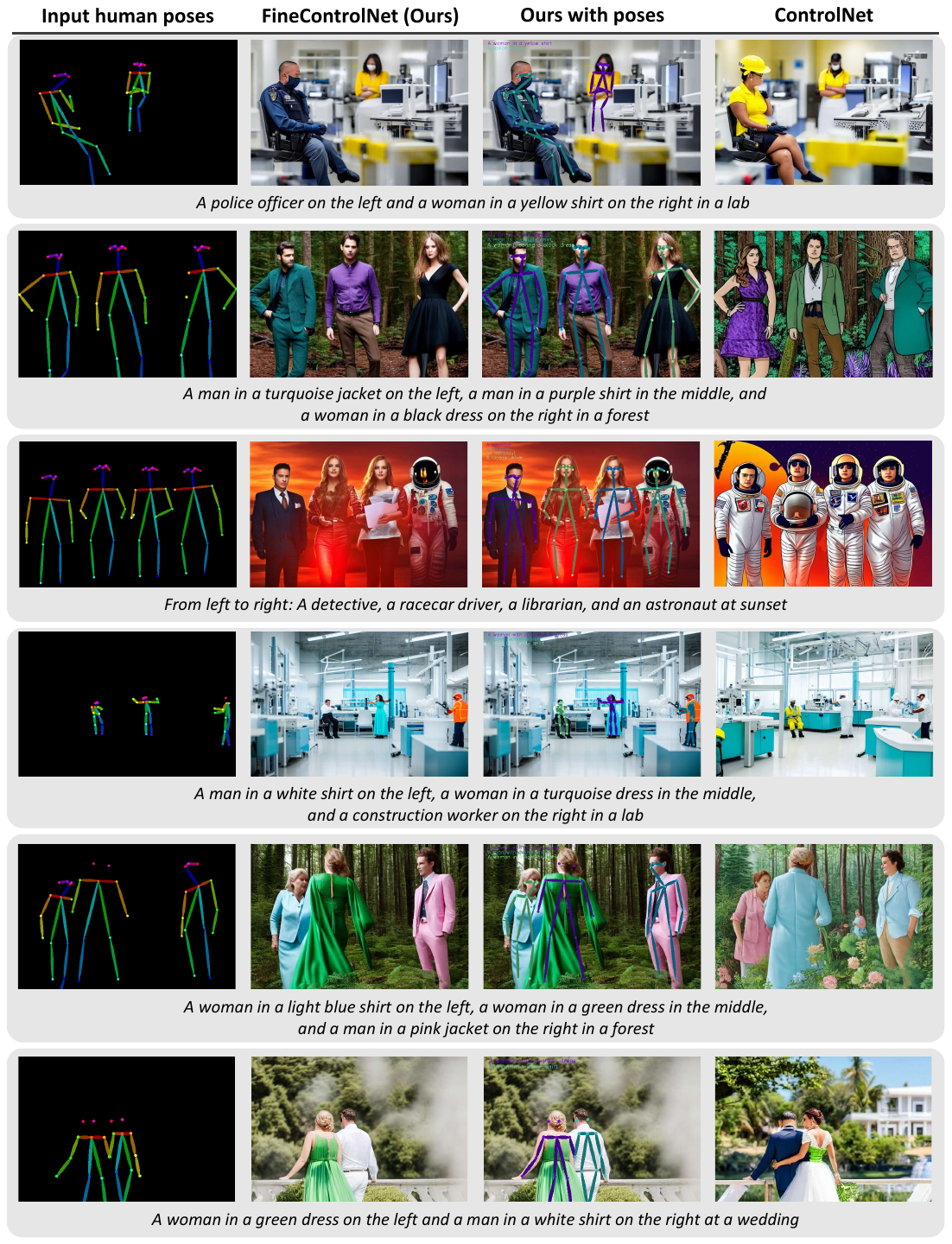}
\end{center}
\vspace{-6mm}
\caption{
\small{
Additional supplementary results demonstrating our method's ability to finely control each instance in the image. We show the input poses (left) and prompt (bottom) along with the results from our method with and without overlaid poses (middle), and ControlNet's~\cite{zhang2023adding} output with the same text prompt (right) for comparison.}
}
\label{fig:more_results1}
\vspace{-2mm}
\end{figure*}

\begin{figure*}[p]
\begin{center}
\includegraphics[width=0.9\linewidth]{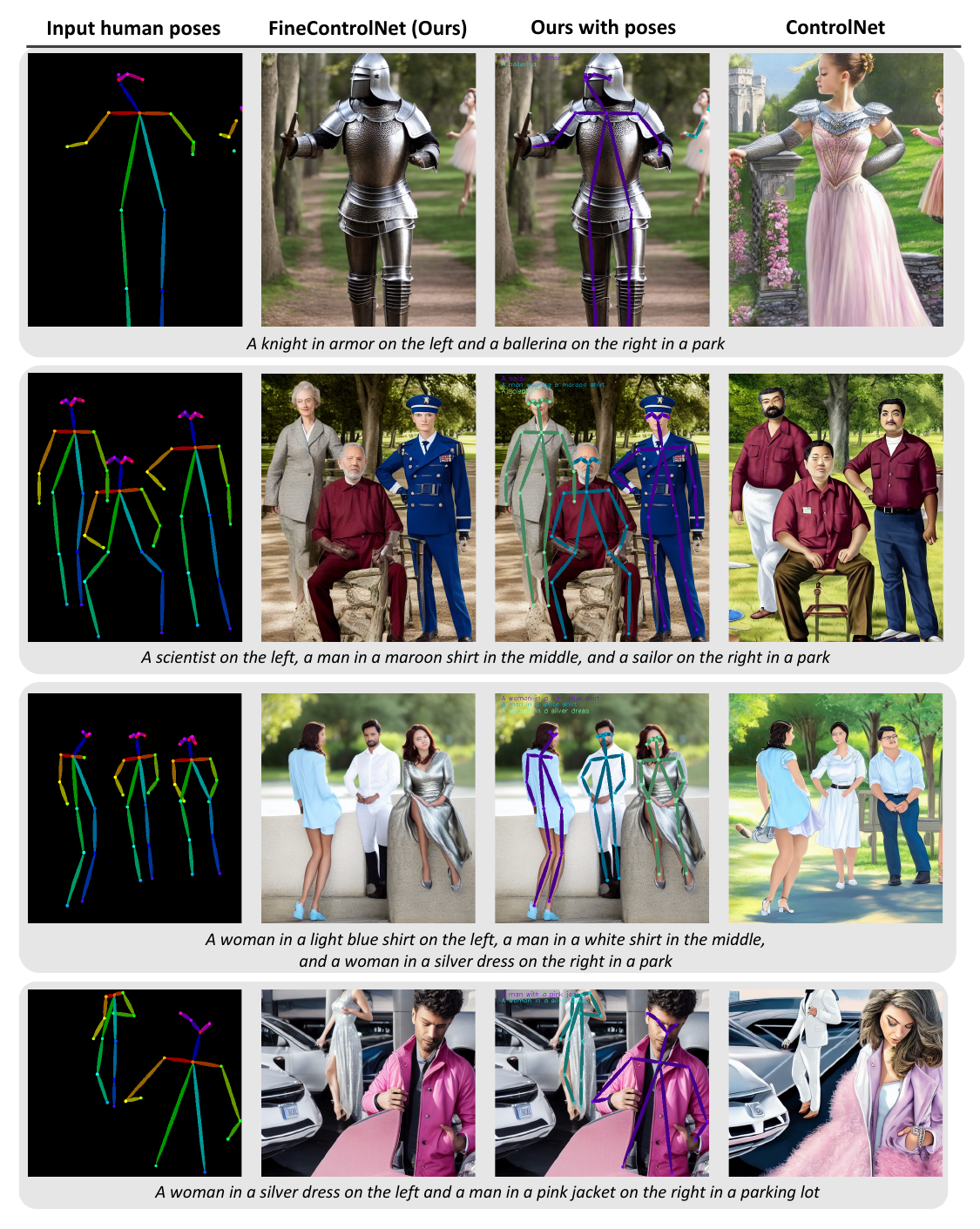}
\end{center}
\vspace{-6mm}
\caption{
\small{
Additional supplementary results demonstrating our method's ability to finely control each instance in the image. We show the input poses (left) and prompt (bottom) along with the results from our method with and without overlaid poses (middle), and ControlNets's~\cite{zhang2023adding} output with the same text prompt (right) for comparison.}
}
\label{fig:more_results2}
\vspace{-2mm}
\end{figure*}

Additional qualitative results of \ourmethod{}'s ability to address instance-specific constraints are shown in Figures~\ref{fig:more_results1} and~\ref{fig:more_results2}. The input poses and prompts are shown in the leftmost columns and at the bottom of each row of images, respectively. The results of \ourmethod{} are provided in the middle two columns, with and without the poses overlaid on the generated images. We also show the outputs of ControlNet~\cite{zhang2023adding} using the same pair of input poses and text prompts as a reference for comparison in the rightmost columns. For both methods, we use the same seed numbers which are sampled from a uniform distribution.

\section{Limitations}
\begin{figure*}[!t]
\begin{center}
\includegraphics[width=1\linewidth]{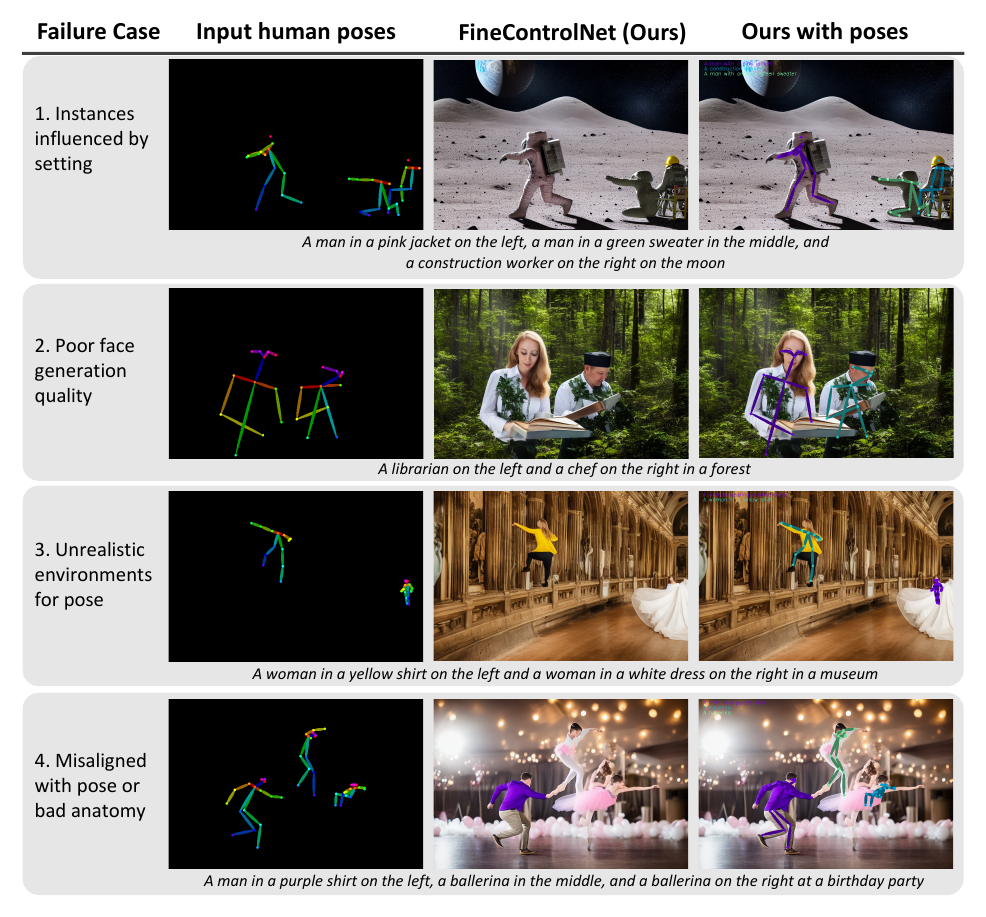}
\end{center}
\vspace{-3mm}
\caption{
\small{
Failure cases. We demonstrate possible failure cases of \ourmethod{} that will be further studied in future work.
}
}
\label{fig:failure_cases}
\vspace{-4mm}
\end{figure*}

Despite showing promising results, our method can sometimes suffer from several failure modes, which include: 1) instance-specific controls being affected by the setting description, 2) human faces synthesized with poor quality, 3) implausible environments for the specified poses, and 4) misaligned input poses and generated images. The results of \ourmethod{} showing these failures are presented in Figure~\ref{fig:failure_cases}.

We observe that instance controls may get altered by the text prompt for the setting, especially in environments with small diversity of instances in the training dataset of images used for Stable Diffusion~\cite{rombach2022high}. In addition, similar to ControlNet~\cite{zhang2023adding}, our method can synthesize human faces that look unrealistic. We also can see unrealistic pairings of instances and environments in some of the generated images by \ourmethod{}. Even when satisfying the instance and setting specifications separately, our method can generate physically implausible scenes, such as floating people, as it does not have an explicit mechanism that prevents from doing so. Finally, \ourmethod{} can generate images whose poses are misaligned or with bad anatomy, particularly when the input poses are challenging.

\section{Dataset}

We provide the histograms of numbers of people per image, person's bounding box resolution per image area ratio, and CrowdIndex~\cite{li2019crowdpose} in \figref{fig:dataset_stats}, for our curated dataset.
CrowdIndex computes the ratio of the number of other persons' joints against the number of each person's joints.
Higher CrowdIndex indicates higher chance of occlusion and interaction between people.
The low resolution ratio and the higher CrowdIndex are related to the difficulty of identity and pose control due to discretization in latent space and ambiguity of instance assignment in attention masks.


\end{document}